\RequirePackage{xcolor}
\documentclass[journal,twoside,web]{ieeecolor}

\usepackage{etoolbox}
\makeatletter
\@ifundefined{color@begingroup}
  {\let\color@begingroup\relax
   \let\color@endgroup\relax}{}
\def\fix@ieeecolor@hbox#1{%
  \hbox{\color@begingroup#1\color@endgroup}}
\patchcmd\@makecaption{\hbox}{\fix@ieeecolor@hbox}{}{}
\patchcmd\@makecaption{\hbox}{\fix@ieeecolor@hbox}{}{}
\makeatother

\usepackage{jsen}
\usepackage{cite}
\usepackage{amsmath,amssymb,amsfonts}
\usepackage{algorithmic}
\usepackage{graphicx}
\usepackage{textcomp}
\usepackage{wrapfig}
\usepackage{bm}
\usepackage{url}

\usepackage{booktabs}
\usepackage{multirow}
\usepackage{float}
\usepackage{stfloats}
\usepackage{comment}

\def\BibTeX{{\rm B\kern-.05em{\sc i\kern-.025em b}\kern-.08em
    T\kern-.1667em\lower.7ex\hbox{E}\kern-.125emX}}

\definecolor{abstractbg}{rgb}{0.89804,0.94510,0.83137}
\begin{document}

\title{Layered e-skin for Shear Sensing}

\author{
Qingzheng Cong,
Alexis W. M. Devillard,
Abu Bakar Dawood,
Xinxin Zhang,
Wen Fan,\\
Neri Niccol\`o Dei,
Cem Suulker,
Kaspar Althoefer,
Etienne Burdet,
and Dandan Zhang
\thanks{This work has been submitted to the IEEE for possible publication.
Copyright may be transferred without notice, after which this version may no longer be accessible.}%
\thanks{This work was supported in part by the EC ERC 101118626 EndoTheranostics and UKRI PALPABLE 101092518 grants.}
\thanks{Cong, Dawood, Dei, Suulker and Althoefer are with Queen Mary University of London, E1 4NS, UK.}
\thanks{Cong, Devillard, Fan, Burdet and D Zhang are or were with The Imperial College of Science, Technology and Medicine, London SW7 2AZ, UK. 
Email: \{e.burdet,d.zhang17\}@imperial.ac.uk}
\thanks{X Zhang is with King's College London, SE1 7EU, UK}
\thanks{Cong, Devillard, and Dawood contributed equally to this work.}}

\IEEEtitleabstractindextext{
\fcolorbox{abstractbg}{abstractbg}{
\begin{minipage}{\textwidth}
\begin{abstract}
\begin{wrapfigure}{r}{0.6\textwidth}
    \vspace{-1.4\baselineskip}
    \centering
    \includegraphics[
        width=\linewidth,
        trim={10bp 35bp 15bp 0bp},
        clip
    ]{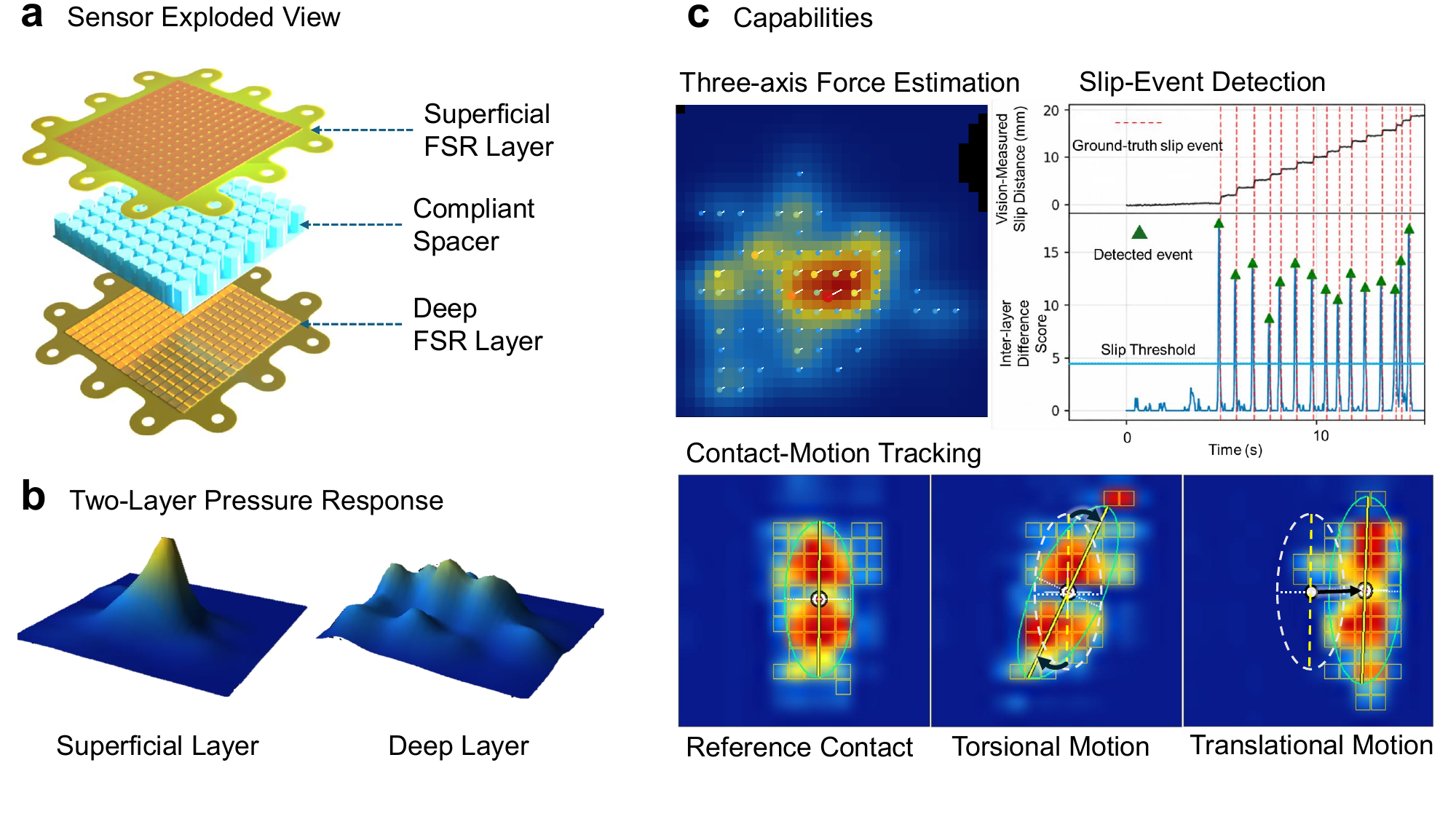}
    \vspace{-25pt}
\end{wrapfigure}
This paper presents a stacked two-layer force-sensing resistor (FSR) array designed for robotic fingertips that combines high-resolution pressure mapping with shear-force estimation. A compliant lattice elastomer spacer converts shear loading into a measurable inter-layer displacement, producing relative center-of-pressure (CoP) shifts between layers. A physics-based moment balance links inter-layer CoP displacement to shear force, while an end-to-end CNN--GRU model captures nonlinear effects from load-dependent compression and contact redistribution. This model, with both layers as input, achieves coefficients of determination R$^2$ =\, 0.914 for F$_x$ and R$^2$ =\, 0.944 for F$_y$, consistently outperforming single-layer baselines for shear-force estimation. Robotic manipulation experiments show that, for contact-motion tracking, the deep layer tracks the translation and rotation imposed by the robot arm, whereas the superficial layer tracks the slip at the contact surface. Transient changes in the difference between the total pressure responses of the two layers provide the best slip-event detection performance among the tested cues. These results demonstrate that two-layer FSR arrays can provide three-axis force estimation, contact-motion tracking, and slip-event detection beyond conventional normal-force sensing.
\end{abstract}

\end{minipage}}}

\maketitle
\thispagestyle{empty}

\section{Introduction}

Tactile sensing is a key enabler of safe, dexterous robotic manipulation. Stable grasping depends not only on normal pressure, but also on friction, shear loading, and incipient slip~\cite{35328b0711884b25ba282f6800d51240, zhang2026slipreview, su2025slipgripper, cong2024tacfr, ss.2021.02}. Distributed tactile arrays are therefore valuable because they can capture the spatial evolution of contact pressure across grasping, sliding, and contact-transition events~\cite{WANG2019111512}.

A wide range of tactile sensing technologies has been developed to provide such contact information~\cite{fan2025crystaltac, MERIBOUT2024115332, lepora2026tactile}. Vision-based tactile sensors (VBTSs), such as GelSight~\cite{yuan2017gelsight}, DIGIT~\cite{lambetaDIGITNovelDesign2020}, and ViTacTip~\cite{fan2024vitactip}, represent a high-resolution class that extracts rich contact information from images. However, the need for imaging hardware, illumination, and optical path length can increase sensor thickness, limiting integration into compact robotic fingertips. Non-vision approaches, such as capacitive~\cite{liSoftCapacitiveTactile2016, dawood2020silicone, dawood2023learning}, magnetic~\cite{bhirangiReSkinVersatileReplaceable2021, magictac}, and multimodal~\cite{fishel2012sensing} tactile sensors, offer compact alternatives for force and deformation sensing. However, they often require precise fabrication, complex electronics, or substantial computational resources. In contrast, \textit{force-sensing resistors} (FSRs) and piezoresistive arrays are thin, low-cost, flexible, and easy to arrange into two-dimensional pressure maps using row-column readout~\cite{frobt, ss.2021.02}. These properties make them attractive for compact robotic fingertips, where sensing area, wiring, and integration complexity are tightly constrained.

Conventional single-layer FSR arrays primarily measure local normal loading and provide limited information about shear forces~\cite{s25103245, https://doi.org/10.1155/2016/9391850, huang2024shearreview}. This is a critical limitation for robotic grasping, where shear force and slip are closely linked to grasp stability. Sensitivity to shear force can be engineered at the structural level, for instance through sandwich structures with dedicated shear-force sensing units~\cite{zhang2024velostat3d} or microstructured elastomer interlayers whose lattice geometry tunes the multiaxial response~\cite{berman2024microlattice}. Another strategy for estimating shear force is to stack pressure-sensitive layers and infer shear force from relative changes between their pressure maps, where Devillard et al.~\cite{eskin} demonstrated that a two-layer FSR skin can estimate pressure direction from inter-layer pressure-map shifts. However, the effective range of shear force that could be estimated remained limited, as the relatively stiff interlayer permitted only small lateral and rotational deformations under typical normal loading. As a result, the stack had only weak shear sensitivity, behaving similarly to two redundant normal-pressure layers.
\begin{figure*}[!t]
    \includegraphics[width=\linewidth, trim=0 110 0 60, clip]{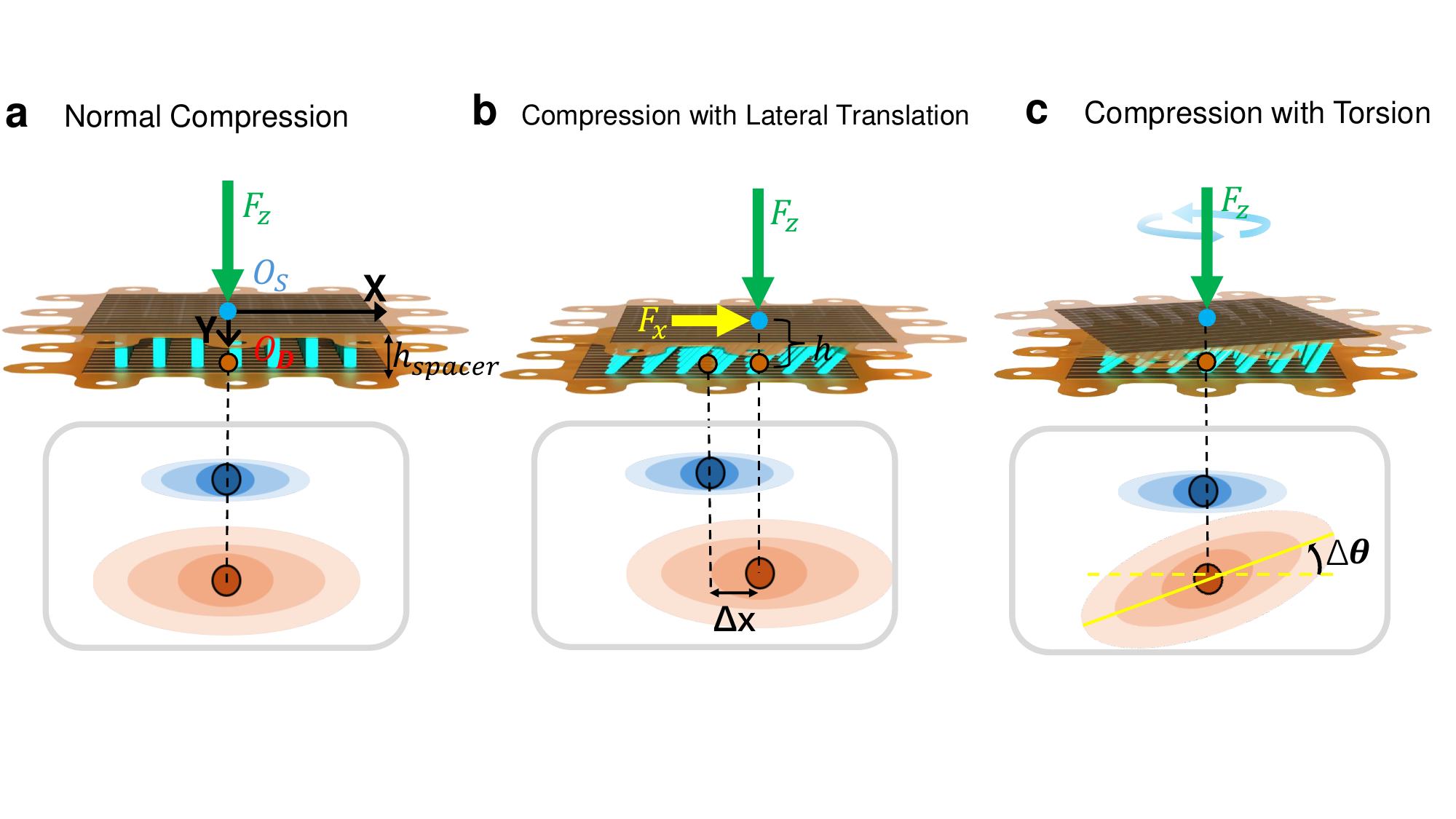}
\caption{Concept of two-layer FSR response under normal, shear, and torsional loading. Blue and red contours represent the pressure distributions on the superficial and deep sensing layers. (a) Under normal compression, the two pressure centers remain aligned; $O_S$ and $O_D$ are the vertically aligned layer origins and $h_{\mathrm{spacer}}$ the undeformed spacer height. (b) Under lateral translation, shear deformation  of the spacer offsets the pressure centers by $\triangle x$; $h$ is the fitted effective height above the deep sensing plane at which the shear force acts. (c) Under torsion, the pressure distribution becomes directionally biased, and its principal axis (PCA) angle $\triangle\theta$ describes the response.}
    \label{fig:main}
    \vspace{-0.7cm}
\end{figure*}

Our idea to address this issue is to introduce a compliant, lattice-structured elastomer spacer between two FSR layers that converts shear loading into measurable inter-layer pressure redistribution, as shown in Fig. \ref{fig:main}. We develop a CNN--GRU model to capture nonlinear deformation and accurately estimate the three-axis force, in line with recent learning-based tactile force estimation~\cite{shahidzadeh2025feelanyforce, shang2026forte}. We also demonstrate that the same two-layer response enables both continuous contact-motion tracking and transient slip-event detection. 

\section{Sensor Fabrication and Force Modeling}
\subsection{Sensor Fabrication and Readout}
 \begin{figure*}[b]
    \centering
    \vspace{-3.5mm}
    \includegraphics[width=\linewidth, trim=0 105 0 130, clip]{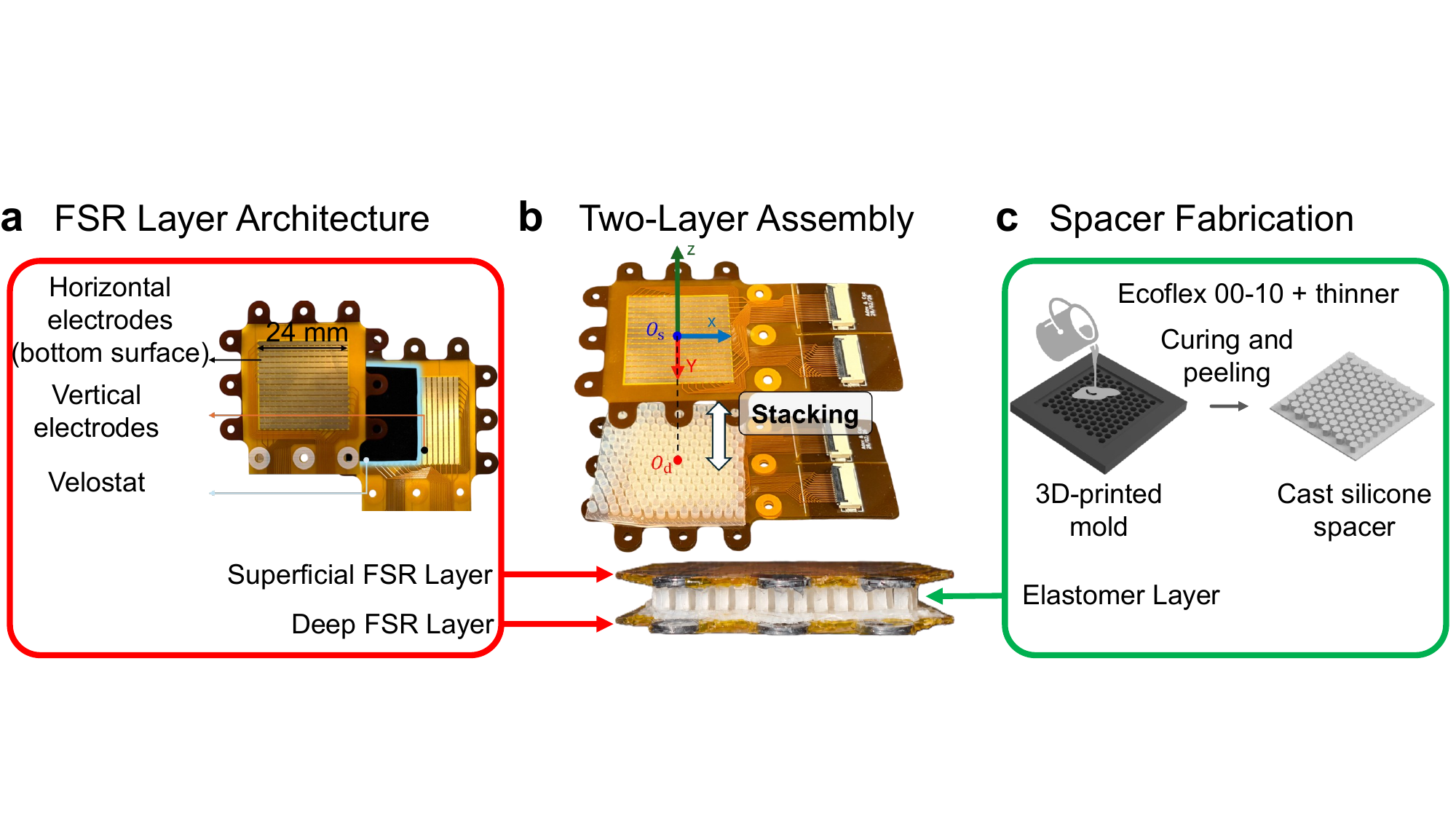}
    \caption{Fabrication of the two-layer FSR array. (a) Single-layer FSR structure formed by laminating a Velostat sheet between crossed FPCB electrodes. Each layer has vertical and horizontal electrodes on opposite faces, and the electrode array is a square with 24\,mm side. (b) Final two-layer sensor assembled by stacking two FSR layers with the compliant spacer in between. (c) Cast Ecoflex 00-10 silicone spacer with a hexagonal lattice structure.}
    \label{fig:fabrication}
     \vspace{-3 mm}
\end{figure*}
The developed e-skin consists of two stacked $16\!\times16$ FSR layers separated by a compliant Ecoflex (Smooth-On, Inc.) silicone spacer as shown in Fig.\,\ref{fig:fabrication}c. Each FSR layer is formed by laminating a Velostat sheet between crossed flexible printed circuit board (FPCB) copper electrodes. The crossed electrodes form a $16\!\times\!16$ taxel array with a pitch of 1.5\,mm, comprising a 1.2\,mm electrode width and a 0.3\,mm gap, resulting in a $24\,$mm $\!\times\!\,24$\,mm sensing area.
The spacer is molded into a 2.5\,mm high hexagonal lattice structure with a 2\,mm feature size, using Ecoflex 00-10 (Smooth-On) diluted with silicone thinner at 10\% by mass. A 0.2\,mm silicone cover layer is added on the contact side for protection and friction enhancement. All silicone components are bonded using silicone adhesive (Sil-Poxy, Smooth-On).

The crossed electrodes are read out via row--column multiplexing. Rows are sequentially selected using 16:1 analog multiplexers (CD74HC4067), while the 16 column signals are measured through voltage dividers and digitized by a 12-bit 16-channel ADC (MAX11632). The readout scans a complete pair of $16\!\times\!16$ frames from the superficial and deep layers in an average of 2.36\,ms, corresponding to a two-layer frame rate of approximately 424\,Hz.

\subsection{Linear Force Estimation}
\label{sec:shear_obs}

As illustrated in Fig.\,\ref{fig:main}, shear loading produces a relative shift between the \textit{superficial} and \textit{deep} layer pressure maps. To quantify this effect, after baseline subtraction and normalization, each layer provides a \textit{pressure map} $\mathbf{P}\!=\!\{p_{i,j}\}$ at every time step. The \textit{center of pressure} (CoP) of each layer is defined as:
\begin{equation}
\mathbf{c} =
\sum_{i,j} 
\frac{p_{i,j}}{\varepsilon + p}
\begin{bmatrix} x_i \\ y_j \end{bmatrix}, \quad p = \sum_{i,j}\, p_{i,j}\,,
\label{eq:cop}
\end{equation}
where $(x_i, y_j)$ are the \textit{taxel coordinates}, $p$ is the \textit{total pressure} over the map, and $\varepsilon$ is a small constant to avoid divergence when the measured pressure is negligible. The CoP displacement between the two layers is then:
\begin{equation}
\triangle\mathbf{c}
  = \, \mathbf{c}^{(s)} - \mathbf{c}^{(d)}
  = \begin{bmatrix} \triangle x \\ \triangle y \end{bmatrix},
\label{eq:delta_cop}
\end{equation}
where superscripts $^{(s)}$ and $^{(d)}$ indicate the superficial and deep layers, respectively. Assuming small deformations and that the superficial-layer CoP approximates the point of load application, a shear force $F_x$ acting at an effective height $h$ above the deep sensing plane shifts the deep-layer CoP in the direction of the applied shear force. The resulting offset between the two layer CoPs allows the normal force ($F_z$) to generate a counteracting moment. Therefore, up to the sign determined by the definition of $\triangle\mathbf{c}$ in Eq.~\ref{eq:delta_cop},
\begin{equation}
F_x\, h \;\approx\; F_z\,\triangle x \, ,
\qquad
F_y\, h \;\approx\; F_z\,\triangle y \,.
\label{eq:moment}
\end{equation}
This expresses that the inter-layer CoP displacement, scaled by the normal load, is physically coupled to the applied shear force.  However, because the compliant spacer undergoes load-dependent compression and nonlinear deformation, the effective height $h$ varies with $F_z$, contact area, and shear direction. Consequently, the following linear regression model is expected to have limited predictive power.

Each force component is regressed using the feature vector
\begin{equation}
\boldsymbol{\phi}
=
\left[
\triangle x,\,
\triangle y,\,
p^{(s)},\,
p^{(d)},\,
\triangle x \, \bar{p},\,
\triangle y \,\bar{p}
\right]^{\mathsf{T}},
\label{eq:linear-features}
\end{equation}
where \(p^{(s)}\) and \(p^{(d)}\) denote the total pressure of the superficial and deep layers, respectively, and \(\bar{p}=[\,p^{(s)}+p^{(d)}]/2\) is their mean. The product terms \(\triangle x \, \bar{p}\) and \(\triangle y \, \bar{p}\) combine inter-layer CoP displacement with a proxy for normal load and capture the first-order relationship suggested by Eq.\,\ref{eq:moment}.

\subsection{Temporal Multi-Head CNN--GRU Model}

To capture nonlinear mappings from data to force, we use an end-to-end model with three output heads, one per force component. After smoothing and downsampling as described in Section \ref{Force Sensing Identification}, the model maps sequences of FSR frames from the selected layer configuration directly to $F_x$, $F_y$, and $F_z$.  To quantify the contribution of each sensing layer, the network is trained using three input configurations: superficial-layer-only, deep-layer-only, and two-layer. Each input sample contains \(T\) consecutive frames. For the superficial-layer-only and deep-layer-only configurations, each frame contains one \(16\!\times\!16\) sensing map. For the two-layer configuration, the superficial-layer map, deep-layer map, and their signed difference are stacked as three input channels. The resulting input-window dimensions are \(T\!\times\!C\!\times\!16\!\times\!16\), where \(C=1\) for either single-layer configuration and \(C=3\) for the two-layer configuration.

The same lightweight CNN encoder is applied independently to all \(T\) frames, producing one 64-dimensional feature vector per frame and therefore a temporal feature sequence of size \(T\!\times\!64\). This sequence is processed by three independent gated recurrent unit (GRU)~\cite{cho2014gru} branches, one for each force component. The complete architecture is summarized in Table~\ref{tab:network_architecture}.

The model was trained using Adam with a learning rate of \(3\times10^{-4}\) and a batch size of 256. A weighted Smooth-\(L_1\) loss was applied to standard-deviation-normalized force errors to balance the three force components and reduce sensitivity to outliers. Training was limited to 200 epochs with early stopping.

\begin{table}[h]
\centering
\caption{Architecture of the  CNN--GRU network. Dimensions are for one sample. All convolutions use $3\times3$ kernels with padding 1. Dropout with probability 0.1 is applied after the fully connected projection.}
\label{tab:network_architecture}
\small
\setlength{\tabcolsep}{4pt}
\renewcommand{\arraystretch}{1.15}
\begin{tabular}{llr}
\toprule
Operation & Configuration & Output shape \\
\midrule
Input window & \(C\in\{1,3\}\) & \(T\!\times\!C\!\times\!16\!\times\!16\) \\
\midrule
\multicolumn{3}{@{}l}{\itshape Shared frame encoder} \\
Conv 1 & \(C\!\rightarrow\!32\), stride 1 & \(T\!\times\!32\!\times\!16\!\times\!16\)\\
Conv 2 & \(32\!\rightarrow\!64\), stride 2 & \(T\!\times\!64\!\times\!8\!\times\!8\) \\
Conv 3 & \(64\!\rightarrow\!64\), stride 1 & \(T\!\times\!64\!\times\!8\!\times\!8\) \\
Flatten & -- & \(T\!\times\!4096\) \\
Projection & \(4096\!\rightarrow\!64\), ReLU & \(T\!\times\!64\) \\
\midrule
\multicolumn{3}{@{}l}{\itshape \(k\in\{x,y,z\}\)} \\
\(\mathrm{GRU}_k\) & hidden 64, 1 layer & \(T\!\times\!64\) \\
Last time step & -- & \(64\) \\
\(\mathrm{Linear}_k\) & \(64\!\rightarrow\!1\) & \(1\) \\
\midrule
Force output & \((\hat{F}_x,\hat{F}_y,\hat{F}_z)\) & \(3\) \\
\bottomrule
\end{tabular}
\end{table}
\section{Experiments}
In all experiments, the superficial FSR layer contacted the environment, whereas the deep FSR layer was fixed to the support or gripper finger beneath the compliant spacer. A \textit{calibration dataset} was first collected under controlled normal and shear loading to fit the force-regression models. A robotic manipulation experiment was then used to test whether the two-layer pressure maps provide spatial cues for contact-motion tracking and slip-event detection in translational and rotational motion.
\subsection{Force Estimation Experiment}

\label{Force Sensing Identification}
Interaction data were collected using a custom three-axis linear platform that applied vertical pressing and in-plane sliding on the e-skin via a 3D-printed rigid hemispherical indenter (radius 2.5\,mm). A six-axis force/torque sensor (ATI Mini40) mounted beneath the e-skin provided synchronized reference forces $F_x$, $F_y$, and $F_z$.

The dataset collected contains 899 contact trials, including 500 normal-pressing trials and 399 shear-sliding trials across 100 contact positions. At each position, five normal-pressing trials were performed by moving the probe vertically along the z-axis. Four shear-sliding trials were then performed from the preloaded contact position in a randomly selected direction in the XY plane, with a random displacement uniformly sampled from $[-0.5,\,0.5]\,\mathrm{mm}$ along the Z-axis. Each sliding motion was terminated when the target shear-force magnitude of 2 N was reached, or the e-skin boundary was encountered, or the normal-force safety limit of 6 N was triggered.

The recorded force ranges were \(F_x \in [-2.01, 2.01]\)\,N, \(F_y \in [-1.98, 1.95]\)\,N, and \(|F_z| \in [0, 6.17]\)\,N. Before training, each contact trial was filtered using a moving average with window size 5, and downsampled by averaging consecutive groups of five samples, yielding 82,513 frames.

Contact trial data were used for training (70\%), validation (15\%) and testing (15\%). Data splits were performed at the trial level to avoid temporal leakage between training and testing data. The validation set was not used for the linear regression, as the least-squares fit involves no tuned hyperparameters.

\subsection{Robotic Manipulation Experiment}
\begin{figure*}[!b]
    \centering
    \vspace{-3 mm}
    \includegraphics[width=\linewidth, trim=0 285 60 0, clip]{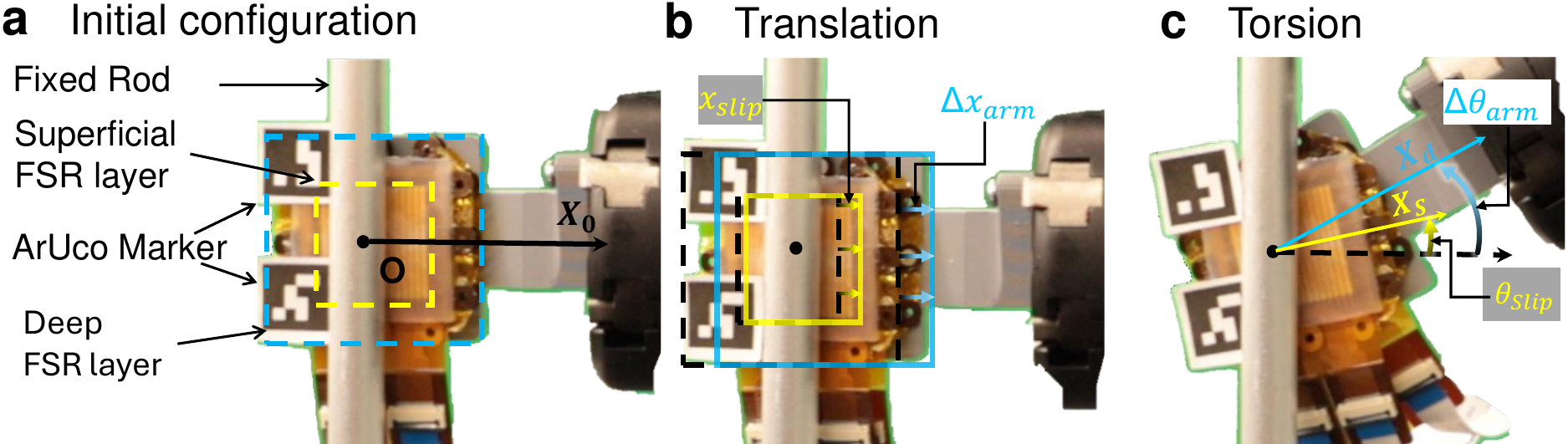}
    \caption{Experimental setup for translational and torsional loading. (a) Initial contact configuration between the two-layer FSR sensor and the fixed rod. The blue dashed box denotes the gripper frame, and the yellow dashed box denotes the contact region of the superficial FSR layer. ArUco markers attached to the superficial layer are used for visual tracking. (b) Translational loading, in which the robot arm translates by $\triangle x_{\mathrm{arm}}$ in $X_0$ direction while the gripper maintains contact with the fixed rod. (c) Torsional loading, in which the robot arm rotates the gripper by an angle $\triangle\theta_{\mathrm{arm}}$. 
    }
    \label{fig:torsion_setup}
    \vspace{-5 mm}
\end{figure*}

Beyond calibrated force estimation, we examine whether the two-layer FSR array can also provide contact-motion tracking and slip-event detection. The experiment was conducted using a UR5e robot arm equipped with a Robotiq Hand-E pressing against a fixed rod with 12 mm in diameter, as shown in Fig.\,\ref{fig:torsion_setup}. The gripper was commanded to approach the rod until the summed FSR readings across both layers exceeded an empirically selected threshold of 500 ADC counts for five consecutive frames. After contact detection, the gripper was commanded to close by an additional distance (g $\in \{6, 9, 12, 15\}$ mm) to increase the normal preload. Here, \(g\), referred to as the grip depth, denotes the commanded post-contact closure rather than the realized finger displacement, sensor indentation, or final jaw gap.

For each grip depth, seven trials were recorded for each motion mode, yielding 28 translational and 28 torsional trials. Each translational trial followed a $0\rightarrow+5\rightarrow0\rightarrow-5\rightarrow0$ mm trajectory along $X_0$ at $2~\mathrm{mm/s}$, whereas each torsional trial followed a $0\rightarrow+25\rightarrow0\rightarrow-25\rightarrow0$ $^{\circ}$ trajectory at $1.5$ $^{\circ}\mathrm{/s}$.

Because an F/T sensor was unavailable in the gripper setup, the mean normal preloads were estimated using the five-frame two-layer CNN–GRU model as $F_z=$\{1.85, 2.82, 3.74, 4.38\}\,N for grip depths \(g=\{6, 9, 12, 15\}\,\mathrm{mm}\), respectively. 

The two layers were evaluated against different references. The deep layer is mechanically fixed to the gripper, which is carried by the robot arm, so its pressure pattern follows the arm motion relative to the rod rather than the slip at the contact. The superficial layer is in contact with the rod and separated from the deep layer by the compliant spacer. When the contact sticks, the spacer shears and the superficial layer stays in place on the rod. Once the contact slips, the superficial layer slides or rotates relative to the rod, and the contact moves across its taxels. Its pressure pattern therefore tracks the slip at the contact surface.

To measure the slip displacement, two ArUco markers~\cite{garrido2014aruco} were attached to a non-contact region of the superficial FSR layer and tracked by an externally fixed Intel RealSense D435 camera, which provided the marker locations at 30 frames per second, temporally aligned with the tactile measurements. Since the rod support and the camera were fixed, the measured marker pose was used as the reference for superficial-layer motion relative to the rod (vision reference). Small elastic deflections of the rod under loading were not separately compensated. For translational slip, the marker displacement was used as the ground-truth slip displacement; for rotational slip, the marker rotation angle was used as the ground-truth slip angle. In parallel, the robot arm's end-effector pose was used as the deep-layer reference (the arm reference), with its lateral displacement and rotation angle representing translation and torsion relative to the fixed rod, respectively.

Continuous contact-motion tracking quantified the accumulated displacement or rotation, whereas slip-event detection identified the timing of individual stick--slip transitions. 

For slip-event detection, the vision-measured translational displacement or rotation angle was filtered using a fourth-order
Butterworth low-pass filter with an 8~Hz cutoff, and differentiated to obtain the corresponding velocity. Following vision-based slip-labeling practice~\cite{zenha2025dense}, ground-truth slip events were defined as local peaks in the velocity magnitude $|v|$. For each trial, peaks were detected using an adaptive threshold of $\mathrm{median}(|v|)+1.4826\times\mathrm{MAD}(|v|)$, and a minimum inter-peak interval of $150$ ms. Here MAD denotes the median absolute deviation and the factor $1.4826$ converts the MAD into a robust estimate of the standard deviation under Gaussian noise.

Each tactile cue was band-pass filtered at \(0.3\)--\(12\)\, Hz, differentiated, converted to a \(60\) ms RMS envelope, and normalized using its median and median absolute deviation. Candidate slip events were extracted from local maxima of the resulting score. Detection thresholds and fixed temporal offsets were selected using training trials only. Detected events were matched one-to-one with the visual events within a tolerance of $\pm100$\,ms. Performance was evaluated using leave-one-trial-out (LOTO) validation. 


\section{Results and Discussion}
\subsection{Sensor Characterization}

Uniformity and repeatability were evaluated from the normal-pressing trials, using the summed positive response of both layers interpolated at a matched normal load. At \(2\,\mathrm{N}\), the response across the 100 contact positions had a coefficient of variation of \(14.5\%\) (bootstrap \(95\%\) CI: \(12.5\)--\(16.2\%\)). Across the 88 positions with a shared loading interval above \(0.3\,\mathrm{N}\) for all five presses, the median within-position coefficient of variation was \(6.2\%\).

The shear trials were used to quantify the inter-layer CoP displacement under shear loading. At shear magnitudes of $1.9$--$2.1$\,N, the trial-wise median inter-layer CoP displacement was $2.20$\,mm ($10$th--$90$th percentile: $1.41$--$2.97$\,mm).
To evaluate the first-order relation in Eq.~\ref{eq:moment}, $F_z\triangle\mathbf{c}$ was fitted to the reference shear force over all shear trials, giving effective coefficients of $h_x=3.64$\,mm and $h_y=4.71$\,mm. Repeating the same fit within normal-force intervals showed that these fitted coefficients increased with normal load, from $1.62/1.44$\,mm for $x/y$ at $0.5$--$1.0$\,N to $4.84/6.53$\,mm at $5.0$--$6.5$\,N. Over the same intervals, the $R^2$ of the regression between $F_z\triangle\mathbf{c}$ and the reference shear force increased from $0.159/0.125$ to $0.626/0.725$ for $x/y$. Thus, \(h_x\) and \(h_y\) represent effective force--CoP coupling coefficients rather than the physical spacer height.

\subsection{Force Estimation}
To assess the contributions of inter-layer and temporal information to force estimation, we compared the physics-inspired linear baseline with CNN--GRU models using different layer inputs and sequence lengths.
\begin{table}[!t]
\centering
\footnotesize
\caption{Per-axis force-estimation performance. Lin.: linear regression using the six features in Eq.~\ref{eq:linear-features}; T/S/D: CNN--GRU with two-layer/superficial/deep inputs; s1/s5: sequence lengths of 1/5 frames. MAE: mean absolute error; best values are in bold.}

\label{tab:multihead_force}
\setlength{\tabcolsep}{3.2pt}
\renewcommand{\arraystretch}{1.05}

\begin{tabular}{ll|c|cc|cc|cc}
\toprule
& Metric & Lin. & T-s1 & T-s5 & S-s1 & S-s5 & D-s1 & D-s5
\tabularnewline
\midrule

\multirow{3}{*}{$F_x$}
& $R^2$ & 0.474 & 0.888 & $\mathbf{0.914}$ & 0.842 & 0.869 & 0.426 & 0.342
\tabularnewline
& MAE (N) & 0.275 & 0.098 & $\mathbf{0.094}$ & 0.126 & 0.118 & 0.217 & 0.231
\tabularnewline
\midrule

\multirow{3}{*}{$F_y$}
& $R^2$ & 0.443 & 0.937 & $\mathbf{0.944}$ & 0.900 & 0.894 & 0.457 & 0.396
\tabularnewline
& MAE (N) & 0.229 & 0.068 & $\mathbf{0.065}$ & 0.085 & 0.087 & 0.175 & 0.178
\tabularnewline
\midrule

\multirow{3}{*}{$F_z$}
& $R^2$ & 0.902 & 0.988 & 0.983 & $\mathbf{0.992}$ & 0.989 & 0.928 & 0.926
\tabularnewline
& MAE (N) & 0.318 & 0.105 & 0.113 & $\mathbf{0.094}$ & 0.104 & 0.252 & 0.256
\tabularnewline

\bottomrule
\end{tabular}
\end{table}

\begin{figure}[!t]
\vspace{-3mm}
\centering
\includegraphics[width=\linewidth, trim=0 30 180 0, clip]{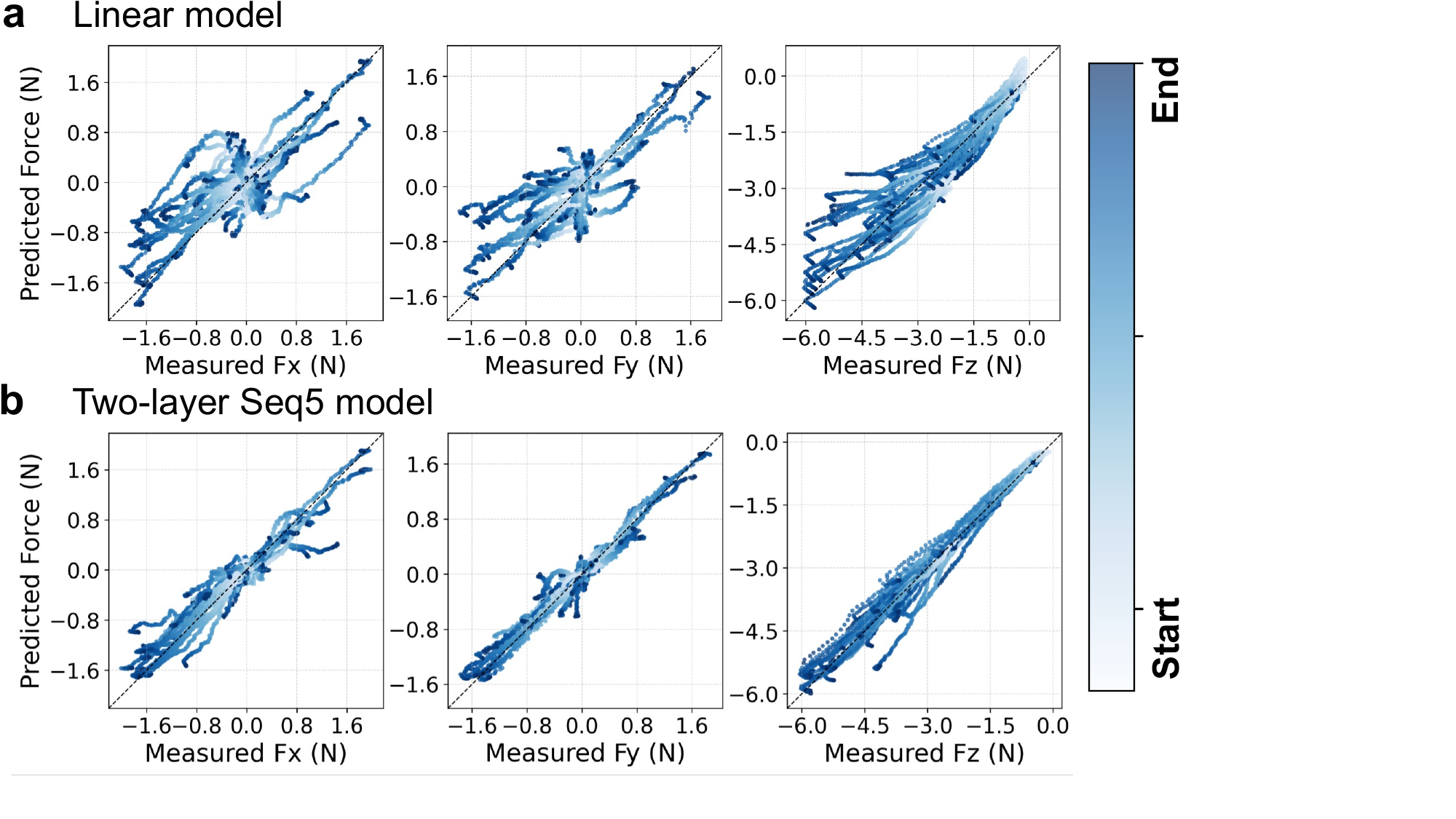}
\caption{Per-axis force estimation from (a) the linear regression and (b) the nonlinear two-layer s5 model. Dashed lines denote the ideal prediction; color indicates temporal progression within each contact trial.}
\label{fig:Regression}
\vspace{-3mm}
\end{figure}

Using both FSR layers as input improves shear-force estimation over using the superficial layer alone. With five-frame inputs (s5), using two layers rather than only the superficial layer increases $R^2$ by 0.045 for $F_x$ and 0.050 for $F_y$, while reducing the corresponding MAEs by 20.3\% and 25.3\%, respectively. The superficial layer alone still carries substantial shear information ($R^2=0.842$--$0.900$), consistent with partial lateral transmission through the spacer. The deep layer alone is far less informative for shear ($R^2=0.342$--$0.457$) but remains effective for  $F_z$  ($R^2\approx0.93$), supporting its role as a mechanically filtered reference that, when combined with the superficial layer, helps isolate inter-layer deformation associated with shear.

The linear regression is substantially less accurate than the two-layer CNN--GRU model. For a sequence length of five (T-s5), the nonlinear model achieves \(R^2=0.914\) and \(0.944\) for \(F_x\) and \(F_y\), respectively, reflecting nonlinear effects (e.g., load-dependent spacer compression, changes in contact area, spatial pressure redistribution, direction-dependent deformation) beyond first order.

Increasing the sequence length from one to five frames improved $F_x$ estimation by a similar amount for the two-layer input ($R^2$ from 0.888 to 0.914) and the superficial-layer input (0.842 to 0.869), but changed $F_y$ only slightly (0.937 to 0.944 and 0.900 to 0.894, respectively). For the deep-layer input, longer sequences reduced accuracy on both shear axes. $F_z$ was already well estimated from a single frame.

Fig.\,\ref{fig:Regression} highlights the limitation of centroid-based features: for the in-plane axes, the linear model follows the trend at large shear but degrades near zero shear because $\triangle\mathbf{c}$ in Eq.~(2) (difference of normalized pressure centroids) becomes noise-sensitive. Operating on full $16\!\times\!16$ pressure maps and a short temporal window, the T-s5 model contracts trajectories towards the diagonal across the full range, lifting $R^2$ for shear force estimation from approximately $0.4$ with the linear model to $0.9$.

\begin{figure*}[!ht]
    \centering
    \includegraphics[width=\textwidth,trim=0 112 10 2, clip]{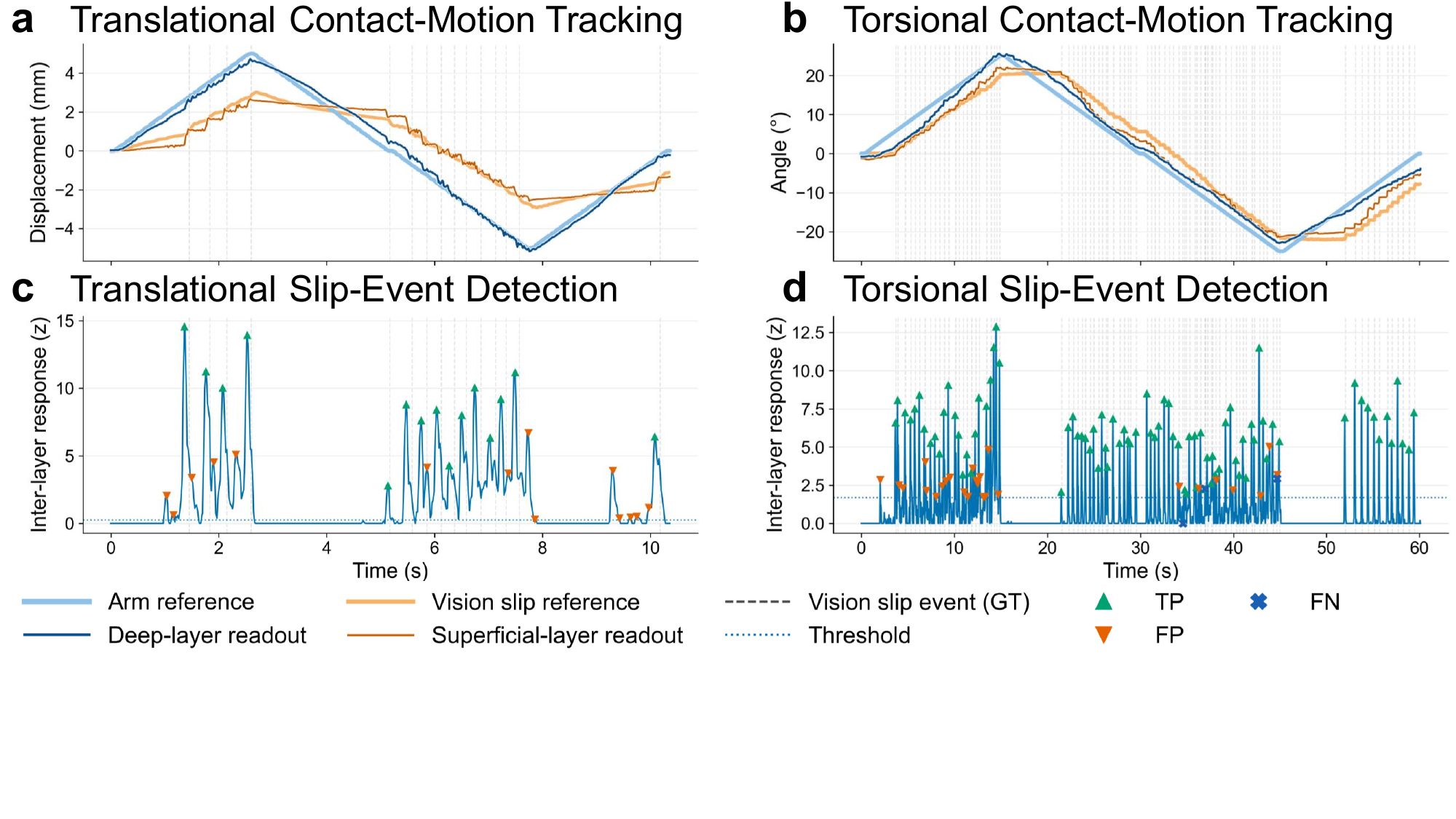}
\caption{Representative layer-specific contact-motion tracking and slip-event detection for translation (a,c) and torsion (b,d) at a grip depth of 12 mm. (a,b) Grip-specific calibrated deep-layer and superficial-layer readouts compared with the arm motion and the vision-measured slip, respectively. (c,d) Slip-event detection using the inter-layer difference cue and the grip-specific threshold learned from the training trials. Vertical dashed lines indicate vision-derived ground-truth slip events. Green triangles, orange inverted triangles, and blue crosses denote true positives (TP), false positives (FP), and false negatives (FN), respectively.}
    \label{fig:slip_detection}
    \vspace{-7mm}
\end{figure*}

\subsection{Layer-Specific Contact-Motion Tracking}
Using the experiments described in Section~III-B, we evaluated contact-motion tracking separately for the deep and superficial layers, against the arm and vision references, respectively. Translation was represented by CoP displacement and torsion by the PCA angle.
\begin{table}[h]
\centering
\caption{Grip-specific LOTO tracking performance. The slope \(a\) is averaged across the LOTO folds, whereas \(R^2\) and MAE are calculated from the pooled out-of-fold predictions. MAE is reported in millimeters for translation and degrees for torsion.}
\label{tab:slip_calib_compact}
\footnotesize
\setlength{\tabcolsep}{2.8pt}
\renewcommand{\arraystretch}{1.04}
\begin{tabular}{l|cccc|ccc}
\toprule
 &  & \multicolumn{3}{c}{Deep (arm)} &
\multicolumn{3}{c}{Superficial (vision)} \\
\cmidrule(lr){3-5}\cmidrule(lr){6-8}
Motion & $g$ & $a$ & $R^2$ & MAE & $a$ & $R^2$ & MAE \\
\midrule
\multirow{4}{*}{Translation}& 6 mm & 1.35 & 0.996 & 0.13 & 0.94 & 0.991 & 0.19 \\
& 9 mm & 1.23 & 0.996 & 0.15 & 0.90 & 0.991 & 0.16 \\
& 12 mm & 1.22 & 0.989 & 0.26 & 0.91 & 0.973 & 0.25 \\
& 15 mm & 1.25 & 0.973 & 0.41 & 0.97 & 0.882 & 0.41 \\
\midrule
\multirow{4}{*}{Torsion}& 6 mm & 1.16 & 0.897 & 3.92 & 0.92 & 0.953 & 2.46 \\
& 9  mm & 1.16 & 0.952 & 2.87 & 0.87 & 0.946 & 2.79 \\
& 12 mm & 1.16 & 0.973 & 2.10 & 0.84 & 0.975 & 1.90 \\
& 15 mm & 1.16 & 0.980 & 1.73 & 0.79 & 0.980 & 1.51 \\
\bottomrule
\end{tabular}
\vspace{ -2mm}
\end{table}
For each grip depth and sensing layer, the readout was calibrated and evaluated using LOTO cross-validation. In each fold, a scale-and-offset calibration,
\(\hat{q}=a q_{\mathrm{FSR}}+b\), was fitted on the remaining trials at the same grip depth and applied to the held-out trial. For translation, \(q_{\mathrm{FSR}}\) is the displacement of the layer CoP (Eq.\,\ref{eq:cop}) relative to its initial position. For torsion, \(q_{\mathrm{FSR}}\) is the PCA angle,
\[
\theta^{(k)}=\tfrac{1}{2}\operatorname{atan2}
\left(2C_{xy}^{(k)},C_{xx}^{(k)}-C_{yy}^{(k)}\right),
\]
where \(C_{xx}^{(k)}\), \(C_{yy}^{(k)}\), and \(C_{xy}^{(k)}\) are the pressure-weighted coordinate covariances of the corresponding superficial or deep layer.

The slope \(a\) represents the scale correction, while \(R^2\) and MAE evaluate the held-out tracking accuracy. Table~\ref{tab:slip_calib_compact} summarizes the results for each grip depth.
\subsubsection{Translation}

Fig.\,\ref{fig:slip_detection}a shows a representative trial at \(g=12\,\mathrm{mm}\). The deep-layer readout followed the imposed arm motion, whereas the superficial-layer readout followed the vision-measured slip, which was smaller than the arm motion because the spacer deformed during sticking.

The deep-layer calibration slope remained within \(a=1.22\)--\(1.25\) at \(g=\{9,12,15\}\,\mathrm{mm}\), but was much higher (\(a=1.35\)) at \(g=6\,\mathrm{mm}\). A likely reason is the weaker deep-layer signal: low-level noise readings spread across the array then carry more weight in the pressure-weighted CoP and pull it toward the array center, which reduces its displacement. By comparison, the superficial-layer slope remained close to unity (\(a=0.90\)--\(0.97\)) across all grip depths. Because this layer is in direct contact with the rod, its CoP follows the contact position and needs only a small scale correction.

Translational tracking accuracy degraded with increasing grip depth for both layers, particularly for the superficial layer. From \(g=6\) to \(15\,\mathrm{mm}\), \(R^2\) fell from \(0.996\) to \(0.973\) for the deep layer and from \(0.991\) to \(0.882\) for the superficial layer, while the MAE rose from \(0.13\) and \(0.19\,\mathrm{mm}\), respectively, to \(0.41\,\mathrm{mm}\) for both. At larger grip depths, deeper indentation and tangential traction during sticking redistribute pressure within the contact, causing the pressure-weighted CoP to shift even when the contact does not slip. Because this traction-dependent shift is not captured by the scale-and-offset calibration, the superficial layer, being closest to the contact, is affected most.

\subsubsection{Torsion}
Fig.\,\ref{fig:slip_detection}b shows a representative trial at \(g=12\,\mathrm{mm}\). Torsional tracking showed the opposite compression dependence, with tracking by both layers generally improving toward \(g=15\,\mathrm{mm}\). At this grip depth, both readouts achieved \(R^2=0.980\), with MAEs of \(1.73^{\circ}\) and \(1.51^{\circ}\), respectively. The deep-layer slope remained near \(1.16\) across all grip depths, suggesting that the PCA angle was less sensitive to limited contact at \(g=6\)\,mm than the translational CoP. This is consistent with the explanation given for translation: low-level noise readings spread evenly across the array have no preferred direction, so they do not rotate the principal axis. They only make the angle noisier, as reflected by the lower \(R^2\) at \(g=6\,\mathrm{mm}\).

As \(g\) increased, the superficial-layer slope decreased from \(0.92\) to \(0.79\), meaning that the pressure pattern rotated 9\% to 27\% more than the slip angle measured by the vision reference. This increasing angular amplification can be explained by two effects. First, since the vision markers measure the motion of the non-contact region whereas the PCA angle describes the pressure distribution within the contact, the two measurements are not always consistent, particularly when the contact pressure is highly non-uniform. Second, under torsion the tangential traction acts in opposite directions at the two ends of the elongated contact, and the resulting pressure shifts rotate the principal axis further than the contact itself, an effect strengthened by deeper indentation. 

A deeper grip therefore affected translation and torsion differently. In translation, it added pressure redistribution that does not scale with slip, so accuracy decreased. In torsion, deeper contact strengthened the pressure pattern, while the associated pressure redistribution was expressed predominantly as a systematic change in angular scale that the grip-specific calibration could absorb. Consequently, torsional tracking accuracy improved with increasing grip depth.
\subsection{Slip-Event Detection}

Slip events were identified from peaks of the vision-measured translational velocity (791 events) and rotational velocity (2,977 events) over 28 trials per motion mode. Five tactile cues were compared: the superficial-layer total response, inter-layer difference (the absolute difference between the superficial-layer and deep-layer total responses), the magnitude of the three-axis force $F_{xyz}$ and the normal force $F_z$, both predicted by the T-s5 model, and a superficial-layer geometric cue (CoP for translation, PCA angle for torsion). For the Global and Grip settings, thresholds were selected within each training fold to maximize F1, and detected events were matched one-to-one with the ground-truth events within $\pm100$ ms. The Oracle setting instead used a label-informed threshold for each held-out trial as an upper bound.

The inter-layer difference achieved the highest F1 score in every setting (Table \ref{tab:slip_event_detection}). A slip transition creates a transient imbalance between the total responses of the two layers, allowing this cue to emphasize differential loading while attenuating changes common to both layers.

\begin{table}[!ht]
\centering
\footnotesize
\vspace{-4 mm}
\caption{Slip-event detection F1 scores (trial-macro, LOTO). Global: one threshold across all grips; Grip: within-grip threshold; Oracle: label-informed per-trial upper bound.}
\label{tab:slip_event_detection}
\setlength{\tabcolsep}{3.0pt}
\renewcommand{\arraystretch}{1.05}
\begin{tabular}{lccc|ccc}
\toprule
& \multicolumn{3}{c}{Translation} & \multicolumn{3}{c}{Torsion} \\
\cmidrule(lr){2-4}\cmidrule(lr){5-7}
Cue & Global & Grip & Oracle & Global & Grip & Oracle \\
\midrule
Inter-layer diff. & $\mathbf{0.672}$ & $\mathbf{0.840}$ & $\mathbf{0.859}$
                  & $\mathbf{0.731}$ & $\mathbf{0.787}$ & $\mathbf{0.809}$ \\
Superficial-layer & 0.642 & 0.795 & 0.836 & 0.621 & 0.681 & 0.730 \\
Geometric cue     & 0.652 & 0.821 & 0.851 & 0.585 & 0.651 & 0.675 \\
$F_{xyz}$         & 0.649 & 0.808 & 0.855 & 0.607 & 0.664 & 0.682 \\
$F_z$             & 0.638 & 0.728 & 0.765 & 0.625 & 0.695 & 0.720 \\
\bottomrule
\end{tabular}
\vspace{- 5mm}
\end{table}

Grip-specific thresholds substantially improved F1 (Translation: $0.672$ to $0.840$; Torsion: $0.731$ to $0.787$),  leaving only \(0.019\) and \(0.022\), respectively, to the oracle bounds. This suggests that the events were generally well separated and that the main limitation of a global threshold was the compression-dependent response amplitude: low preload produced weaker tactile transients, whereas high preload produced larger ones. A fixed latency compensation reduced the median tactile lead relative to vision from $82$\,ms to $10$\,ms, consistent with camera pipeline delay~\cite{realsense_latency}. Representative detections are shown in Fig.\,\ref{fig:slip_detection} c and d.

\section{Conclusion}
This paper presented a two-layer FSR e-skin that extends normal-force sensing to three-axis force estimation by exploiting the CoP displacement induced by shear loading through a compliant lattice spacer. Compared with the stiffer inter-layer design in~\cite{eskin}, the proposed spacer enables shear-force estimation under lower normal preload ($\pm 2$\,N shear force at approximately 6\,N normal load) by allowing larger relative displacement between the two FSR layers.

A first-order moment-balance analysis and related linear regression were consistent with this coupling, while a CNN--GRU model captured the remaining nonlinear response and achieved $R^2=0.914$ and $0.944$ for $F_x$ and $F_y$, respectively. Relative to a superficial-layer-only baseline, the two-layer input increased the mean shear $R^2$ by approximately 0.048 and reduced the mean shear MAE by $22.8\%$.

The robotic experiments further demonstrated that the two layers provide complementary information for contact-motion tracking. The deep-layer CoP and PCA features tracked the imposed arm motion, whereas the superficial-layer features tracked the slip at the contact surface. Transient inter-layer difference also provided a consistent cue for both translational and torsional slip-event detection.

A current limitation is that the relation between spacer compliance and the range of shear force that can be estimated has not yet been systematically characterized. Establishing this mapping across spacer materials and lattice geometries is a necessary next step toward tuning the force range and sensitivity of the two-layer FSR e-skin.
\section*{Acknowledgment}
The authors used OpenAI ChatGPT to assist with English-language editing,
sentence rephrasing, and stylistic refinement throughout the manuscript. All technical
content and interpretations were reviewed and verified by the authors.

\normalsize



\bibliographystyle{IEEEtran}
\bibliography{refs}
\end{document}